\documentclass{article}
\usepackage{spconf,amsmath,epsfig}
\usepackage{algorithm,amsfonts}
\usepackage{subfigure}
\usepackage{graphicx}
\usepackage{makecell}
\begin{document}

\title{Hyperspectral and LiDAR data classification based on linear self-attention}

\name{Min Feng$^{1,2}$, Feng Gao$^{1,2,*}$, Jian Fang$^{1,2}$, Junyu Dong$^{1,2}$}
\address{$^1$College of Information Science and Engineering, Ocean University of China \\
$^2$ Institute of Marine Development, Ocean University of China
\thanks{This work was supported in part by the National Key Research and Development Program of China under Grant 2018AAA0100602, in part by the
National Natural Science Foundation of China under Grant U1706218, and in
part by the Key Research and Development Program of Shandong Province
under Grant 2019GHY112048. (Email: gaofeng@ouc.edu.cn)}}
\maketitle

\begin{abstract}

In this paper,  an efficient linear self-attention fusion model is proposed for the task of hyperspectral image (HSI) and LiDAR data joint classification. The proposed method is comprised of a feature extraction module, an attention module, and a fusion module. The attention module is a plug-and-play linear self-attention module that can be extensively used in any model. The proposed model has achieved the overall accuracy of 95.40\% on the Houston dataset. The experimental results demonstrate the superiority of the proposed method over other state-of-the-art models.

\end{abstract}

\begin{keywords}
hyperspectral image, LiDAR, cross-modal data fusion, classification.
\end{keywords}

\section{Introduction}

Recently, with the development of remote sensing instruments and advanced spectral imaging technology, hyperpsectral images (HSIs) have attracted increasing attention. HSI contains vast amounts of spectral information, which can be used for ground object classification. Nevertheless, HSI alone can hardly yield promising classification results \cite{c}. Researchers begin to seek other powerful approaches and use LiDAR data to compensate for the shortcomings of HSI sensors. Consequently, LiDAR sensors are capable of measuring distances by illuminating the target with laser light. Thereby, they are commonly used to make 3-D representations of the ground. It has been proved that the combination of HSI and LiDAR data can improve the ground object classification. Therefore, joint classification by using HSI and LiDAR data has become a research hotspot in remote sensing communities.

\begin{figure}[h]
\begin{center}
\includegraphics [width=3.4in]{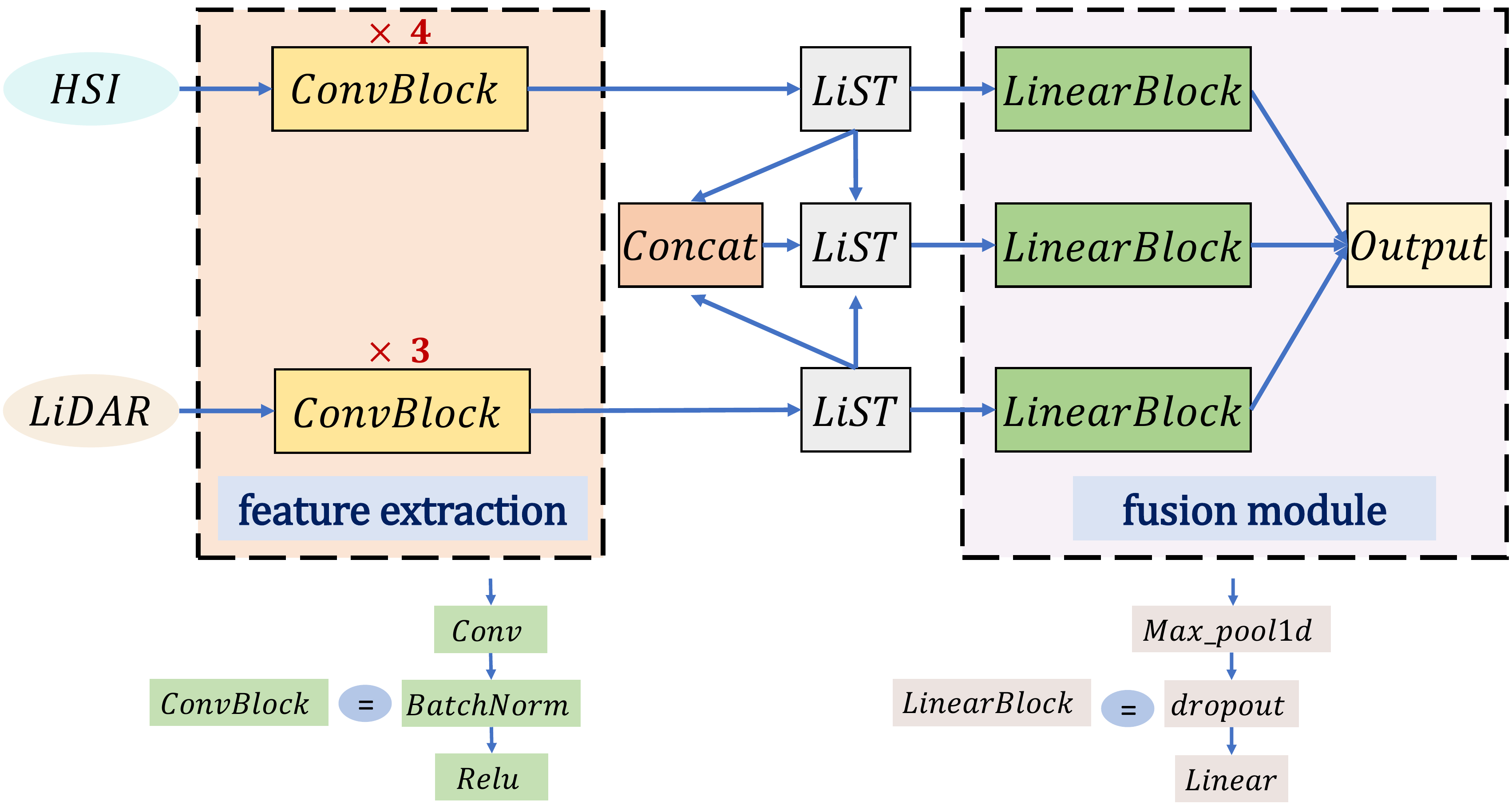}
\caption{Flowchart of the proposed Linear Self-Attention Fusion (LSAF) model.}
\label{model}
\end{center}
\end{figure}

In fact, great efforts have been made to solve the HSI and LiDAR joint classification issue. Debes et al. \cite{d} putted more emphasis on the spatial context, and proposed an approach to extract features from LiDAR data and HSI, respectively. The method achieved the first place in the 2013 GRSS Data Fusion Contest. Liao et al. \cite{a} presented a generalized graph-based fusion method to model the similarities in spatial, spectral, and elevation characteristics of the fused nodes by employing weighted edges. Gu et al. \cite{e} constructed a similarity matrix by exploiting Gaussian kernel functions, which employ different scales to discriminate the classification between heterogeneous features from HSI and LiDAR data.

Recently, numerous studies have used the attention mechanism to improve the classification performance in computer vision tasks. These attention mechanisms have the capability of improving feature representation by explicitly building dependencies among multisource data. The intuition behind the attention mechanism is to train the network to have the ability to learn where to attend and focus on the meaningful parts. In \cite{g}, the attention mechanism was used on HSI and LiDAR data fusion. Meanwhile, Li et al. \cite{h} developed a multiscale residual attention model for HSI and LiDAR to facilitate subsequent feature representation. Although some correlations and indwelling patterns have been captured by existing attention-based methods, they can hardly adaptively adopt the context-aware representations and bridge the heterogeneous gap between HSI and LiDAR data.

In this paper, we propose a novel Linear Self-Attention Fusion model, LSAF for short, which exploits the context-aware representation between HSI and LiDAR data. To be more specific, as illustrated in Fig. \ref{model}, CNN-based feature extraction is first implemented for HSI and LiDAR data, respectively. Afterwards, we introduce a plug-and-play linear self-attention module to enhance the representation of HSI and LiDAR data. Ultimately, we devise a self-adaptive decision fusion module to aggregate the classification results. Experimental results on the Houston dataset demonstrate the superiority of the proposed LSAF model as compared to closely related methods.

\section{Methodology}

As showed in Fig. \ref{model}, the framework of the proposed LSAF is comprised of three components: the feature extraction module, the linear self-attention module, and the self-adaptive decision fusion module. These modules will be described in detail in the remainder of this section.

\subsection{Feature Extraction}

Recently, CNN-based methods have been extremely popular in image feature extraction. For convenience, in this paper, we use the retrained HybridSN model \cite{hybrid} as the feature extractor for hyperspectral images. After each convolutional layer, we apply batch normalization and ReLU activation function. Besides, the LiDAR data is treated in a similar way. Unlike HSI feature extractor, three ConvBlocks are used for LiDAR feature extraction. It should be noted that the feature maps of LiDAR data are consistent in size and dimensions with that of HSI. 

\subsection{Linear Self-Attention Module}

\begin{figure}[h]
\begin{center}
\includegraphics [width=3in]{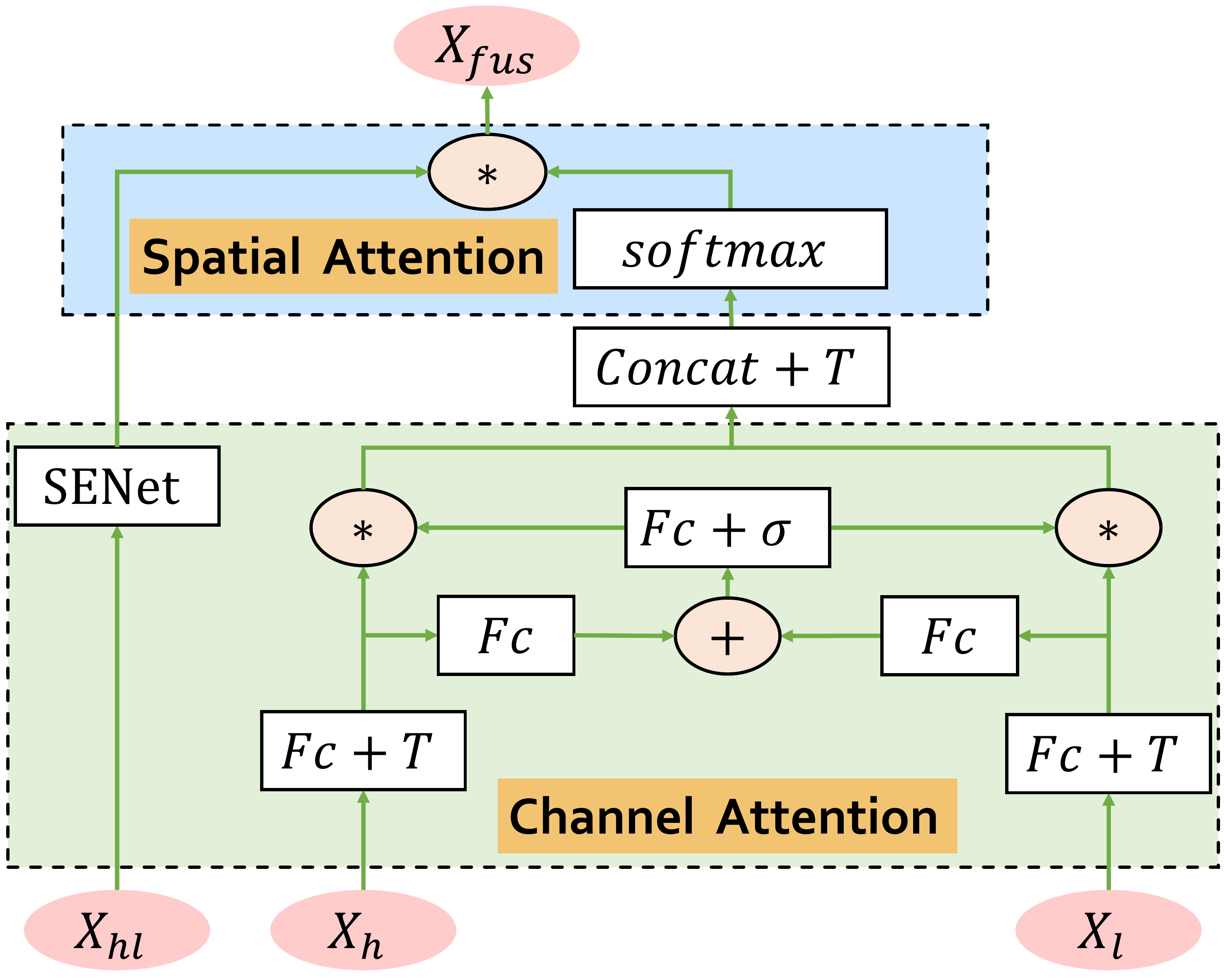}
\caption{Flowchart of the Linear Self-Attention Module (LiST)}
\label{atten}
\end{center}
\end{figure}

Ideally, the attention mechanism is deemed to highlight the relevant features that are conducive to the images. Therefore, a linear self-attention (LiST) module is proposed, as illustrated in Fig. \ref{atten}. Let $X_{h} \in \mathbb{R}^{c \times hw}$ denotes the HSI feature, $X_{l} \in \mathbb{R}^{c\times hw}$ denotes the LiDAR feature, and $X_{hl} \in \mathbb{R}^{2c \times hw}$ denote the concatenated HSI and LiDAR features, respectively. Here $c$ represents the channel dimension, $h$ and $w$ refer to the height and width of the original feature map. To enhance the discriminative power of the three features, a FC layer is employed for feature transformation. From Fig. \ref{atten}, we can see that the LiST module is a fused attention based on a channel attention and spatial attention.

\subsubsection{The Channel-Wise Attention}

SENet \cite{senet} enables the deep network to perform dynamic channel-wise feature recalibration and has achieved significant performance improvement in computer vision tasks. Hence, in this work, we take channel-wise attention into account, and it is imposed to $X_{hl}$. Besides, $\hat{X}_{h}$ and $\hat{X}_{l}$ also follow the similar attention structure after transpose transformation. The channel attention is formally defined as
\begin{equation}
X'_{h} = \sigma(\textrm{FC}(\textrm{FC}(\hat{X}_{h}^{T}) + \textrm{FC}(\hat{X}_{ l}^{T})))* \hat{X}_{h}^{T},
\end{equation}
\begin{equation}
X'_{l} = \sigma(\textrm{FC}(\textrm{FC}(\hat{X}_{h}^{T}) + \textrm{FC}(\hat{X}_{l}^{T}))) * \hat{X}_{l}^{T},
\end{equation}
where $X'_{h} \in \mathbb{R}^{hw \times c}$ and $X'_{l} \in \mathbb{R}^{hw \times c}$ refer to the output of $X_{h}$ and $X_{l}$ after channel attention, $\sigma(\cdot)$ refers to sigmoid function, FC refer to fully-connected layers which are applied to map the channel features, and the primary purpose of summation ($+$) is to achieve feature interaction between HSI and LiDAR data.

Thereafter, we successively concatenate ($\oplus$) and transpose ($T$) $X'_h$ and $X'_l$ to obtain $X'_{fus} \in \mathbb{R}^{2c\times hw}$, which is represented as in Eq. (\ref{Xc}):
\begin{equation}
X'_{fus} = (X'_{h} \oplus  X'_{l})^{T}.
\label{Xc}
\end{equation}

Similarly, $X'_{hl} \in \mathbb{R}^{2c \times hw}$, the feature obtained after SENet, is represented as follows: (\ref{Xcfusion})
\begin{equation}
X'_{hl} = f_\textrm{se}(X_{hl})
\label{Xcfusion}
\end{equation}
\subsubsection{The Spatial Attention}

Considering the spatial weight, the intricate $X'_{fus}$ after the channel attention operation can be applied by the $softmax$ operator. The explicit spatial attention equation is computed as follows:
\begin{equation}
X_{fus} = X'_{hl} * softmax(X'_{fus}).
\end{equation}

\begin{figure*}[htb]
\begin{center}
\includegraphics [width=4.5in]{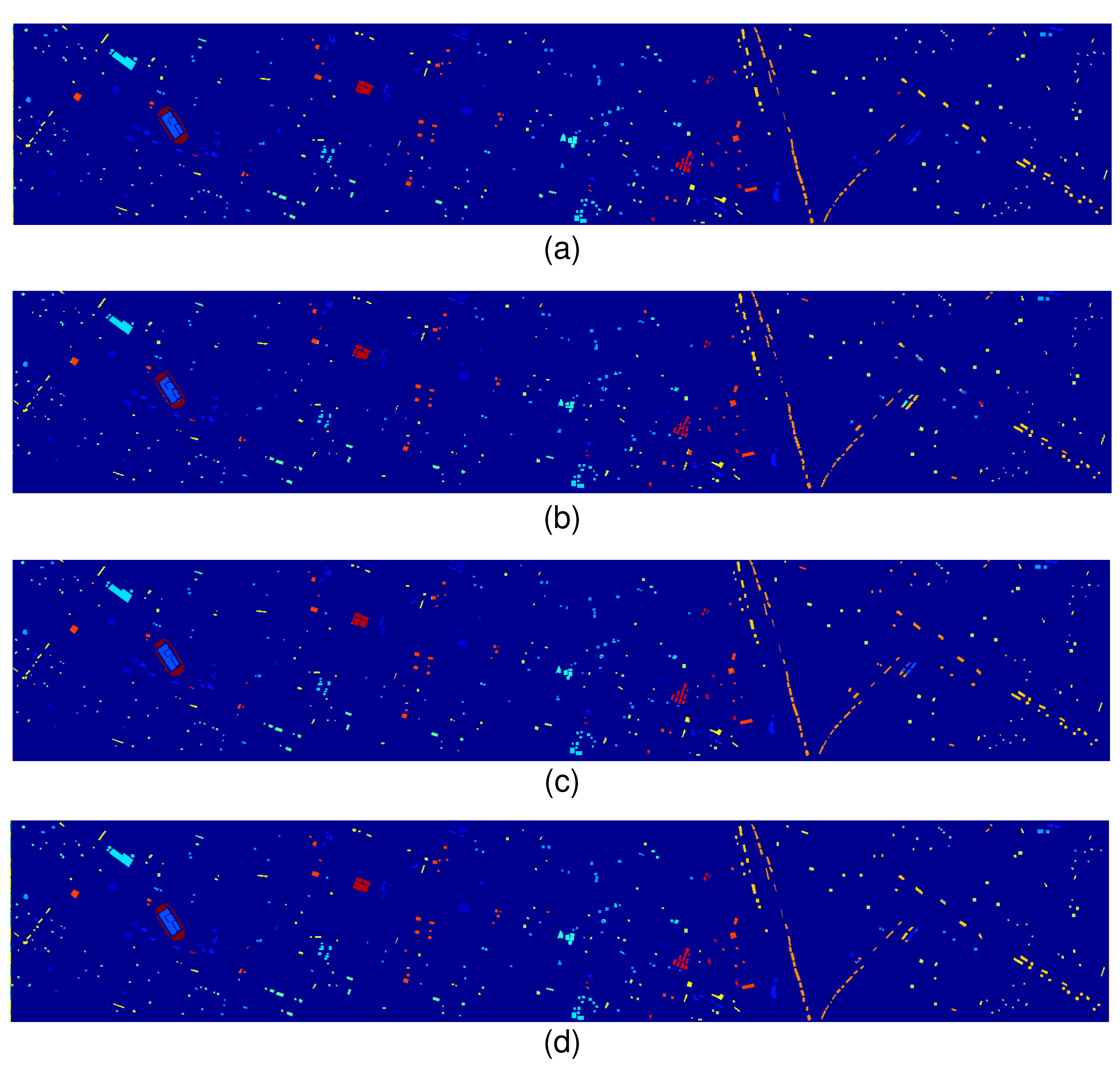}
\caption{Visualization of results of different methods. (a) The ground truth. (b) Result by $A^{3}$CLNN (90.55\%); (c) Result by PToP CNN (92.48\%); (d) Result of the proposed LSAF (95.40\%).}
\label{result}
\end{center}
\end{figure*}

\subsection{The Self-adaptive Decision Fusion Module}

As illustrated in Fig. \ref{model}, we can obtain $Y_{h}$, $Y_{l}$ and $Y_{fus}$, which are the results of $X_{h}$, $X_{l}$ and $X_{fus}$ after LinearBlock. The fusion result $Y$ is a weighted summation of the three outputs as follows: 
\begin{equation}
Y = \lambda_{1} * Y_{h} + \lambda_{2} * Y_{l} + Y_{fus},
\end{equation}
The parameters ${\lambda_{1}}$ and ${\lambda_{2}}$ are obtained by learning in training.

\section{Experimental Results and Analysis}

\subsection{Experimental Data}

To verify the effectiveness of the proposed model, extensive experiments are carried out on the Houston dataset, which is introduced in GRSS Data Fusion Contest 2013. The dataset was captured in 2012 by the Compact Airborne Spectrographic Imager (CASI) sensor over the University of Houston, TX, USA.  Its size is 349$\times$1905 pixels, with a spatial resolution of 2.5 m. These data, together with the reference classes, are available online from the IEEE GRSS website: http://dase.grss-ieee.org/.

The learning rate of the proposed model is set to 0.0001 and the number of epochs is set to 110. The mini-batch size is 128 and Adam is employed as the optimizer. Furthermore, considering the redundant information presented in the original HSI data, principal component analysis (PCA) is employed for spectral dimensionality reduction from 144 to 30 in our experiments.

\subsection{Results and Analysis}

In this paper, we compare the proposed model with two state-of-the-art models, $A^{3}$CLNN \cite{h} and PToP CNN \cite{ptop}. The OA values of all methods are presented in Table \ref{table_res}. The visualized classification results are illustrated in Fig. \ref{result}. As it can be observed that the proposed LSAF model is superior to other methods.

\begin{table}[htbp]
\centering
\renewcommand\arraystretch{1.4}
\caption{Classification performance (\%) using A³CLNN, PToP CNN and proposed LSAF}	
\label{table_res}
\begin{tabular}{c|cccc}
\hline
~~Class~~  & ~~A³CLNN~~ & ~PToP CNN~ & ~LSAF~ \\
\hline \hline
Healthy grass	 & 81.73	& 85.77	& 98.22\\
Stressed grass & 	84.43	& 87.08	& 96.12 \\
Synthetic grass 	& 91.49	& 99.57	& 100.00 \\
Trees	 	& 96.72	 & 94.13	& 95.08 \\
Soil	 & 99.97	& 100.00	 & 96.99 \\
Water		& 97.90	& 99.38 & 	98.92 \\
Residential	 & 87.06	& 87.38 & 	95.54 \\
Commercial	 & 96.93	& 97.35	 & 96.66 \\
Road	 & 87.88	& 90.81 & 	95.50 \\
Highway	& 	70.82 & 	72.21 & 	83.48 \\
Railway	 & 98.13	& 100.00 & 	94.52 \\
Parking Lot 1	 & 94.65	& 98.13	 & 96.90 \\
Parking Lot 2	 & 96.02	& 92.11	 & 99.43 \\
Tennis Court	 & 97.30	& 99.30 & 	100.00 \\
Running Track	 & 96.05	& 100.00 & 	99.79 \\
\hline \hline
OA	 & 90.55	& 92.48 & 95.40 \\
\hline
\end{tabular}
\end{table}

Meanwhile, it can be distinctly illustrated from this table that the proposed LSAF achieves optimal classification results in the categories of healthy grass, stressed grass, road, and parking Lot 2, especially in the case of healthy grass and road, which are two general categories with inferior classification results. We also visualized the classification result graph, as shown in Fig. \ref{result}, where (d) represents the classification result graph obtained by the proposed LSAF, which is the most similar to the human-labeled ground truth from Houston dataset.

\section{Conclusion}

In this paper, we propose a LASF model to solve the problem of HSI and LiDAR joint classification, which is comprised of a feature extraction module, an attention module, and a fusion module. The attention module is a plug-and-play linear self-attention module that can be extensively used in any model. The comparison with the state-of-the-art methods demonstrates the excellent performance of the proposed model. Finally, the optimal accuracy of 95.65\% and the average accuracy of 95.40\% are obtained on the Houston dataset.

\end{document}